\documentclass[conference]{IEEEtran}
\IEEEoverridecommandlockouts
\usepackage{cite}
\usepackage{amsmath,amssymb,amsfonts}
\usepackage{algorithmic}
\usepackage{graphicx}
\usepackage
{textcomp}
\usepackage{xcolor}
\def\BibTeX{{\rm B\kern-.05em{\sc i\kern-.025em b}\kern-.08em
    T\kern-.1667em\lower.7ex\hbox{E}\kern-.125emX}}

\usepackage{graphics}
\usepackage{url}
\usepackage{multirow}
\usepackage{siunitx}
\usepackage{bm}

\usepackage{adjustbox}
\usepackage{diagbox}

\usepackage[switch]{lineno}

\usepackage{color}
\def\red#1{\textcolor{red}{#1}}
\def\blue#1{\textcolor{blue}{#1}}

\begin{document}

\title{Burst Super-Resolution with Diffusion Models for Improving Perceptual Quality
\vspace*{-2mm}
}

\author{\IEEEauthorblockN{Anonymous Authors}}
\author{\IEEEauthorblockN{Kyotaro Tokoro}
\IEEEauthorblockA{\textit{Toyota Technological Institute}, Japan \\
sd24439@toyota-ti.ac.jp}
\and
\IEEEauthorblockN{Kazutoshi Akita}
\IEEEauthorblockA{\textit{Toyota Technological Institute}, Japan \\
sd21501@toyota-ti.ac.jp}
\and
\IEEEauthorblockN{Norimichi Ukita}
\IEEEauthorblockA{\textit{Toyota Technological Institute}, Japan \\
ukita@toyota-ti.ac.jp}
}

\maketitle

\vspace*{-10mm}

\begin{abstract}
While burst LR images are useful for improving the SR image quality compared with a single LR image, prior SR networks accepting the burst LR images are trained in a deterministic manner, which is known to produce a blurry SR image.
In addition, it is difficult to perfectly align the burst LR images, making the SR image more blurry.
Since such blurry images are perceptually degraded, we aim to reconstruct the sharp high-fidelity boundaries.
Such high-fidelity images can be reconstructed by diffusion models.
However, prior SR methods using the diffusion model are not properly optimized for the burst SR task.
Specifically, the reverse process starting from a random sample is not optimized for image enhancement and restoration methods, including burst SR.
In our proposed method, on the other hand, burst LR features are used to reconstruct the initial burst SR image that is fed into an intermediate step in the diffusion model.
This reverse process from the intermediate step 1) skips diffusion steps for reconstructing the global structure of the image and 2) focuses on steps for refining detailed textures.
Our experimental results demonstrate that our method can improve the scores of the perceptual quality metrics.
Code: \url{https://github.com/placerkyo/BSRD}.
\end{abstract}



\section{Introduction}
\label{section:introduction}

Super-Resolution (SR) is a low-level vision task for super-resolving a Low-Resolution (LR) image to its High-Resolution (HR) image.
SR is widely applicable to many real-world tasks, such as image enhancement on consumer cameras and remote sensing.
SR for super-resolving an LR image is called Single-Image Super-Resolution (SISR)~\cite{DBLP:journals/pami/DongLHT16,ntire2018,DBLP:conf/cvpr/HarisSU18,DBLP:conf/iccvw/GuLZXYZYSTDLDLG19,DBLP:journals/pami/HarisSU21}.
However, SISR is not an easy task due to its ill-posed nature.
In particular, SR for a degraded LR image is much more difficult in SISR.
Such degraded images are often captured by, for example, commercially available smartphones.

To improve the SR quality, SR can accept multiple LR images.
Assume that a quasi-static scene is captured many times.
Such a burst shot allows us to capture many burst images with subtle differences, such as sub-pixel displacements and small noise.
By integrating all the burst images into one~\cite{DBLP:conf/cvpr/BhatDGT21}, the SR quality can be improved compared with SISR.
This type of SR is called burst SR~\cite{DBLP:journals/corr/abs-2106-03839,DBLP:conf/cvpr/BhatDTCCCCCDFGG22}.
Since the burst shot is allowed in recent smartphones, burst SR is a promising technique.

\begin{figure}[t]
  \vspace*{-7mm}
  \begin{center}
  \includegraphics[width=\linewidth]{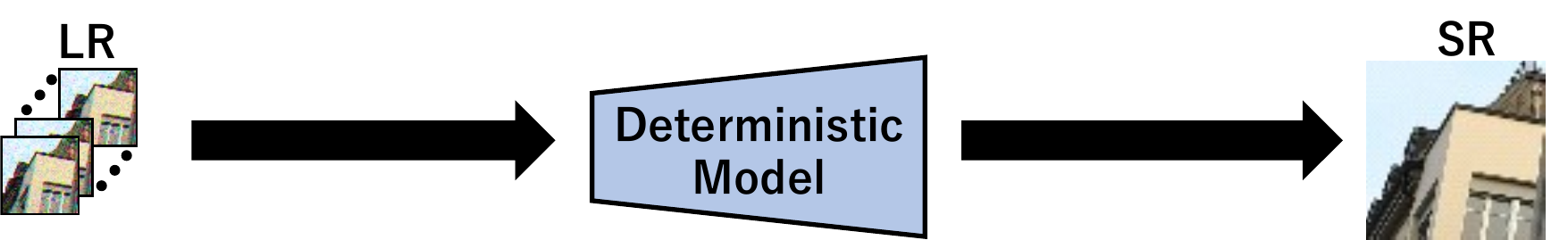}\\
  (a) Deterministic models\\
  \includegraphics[width=\linewidth]{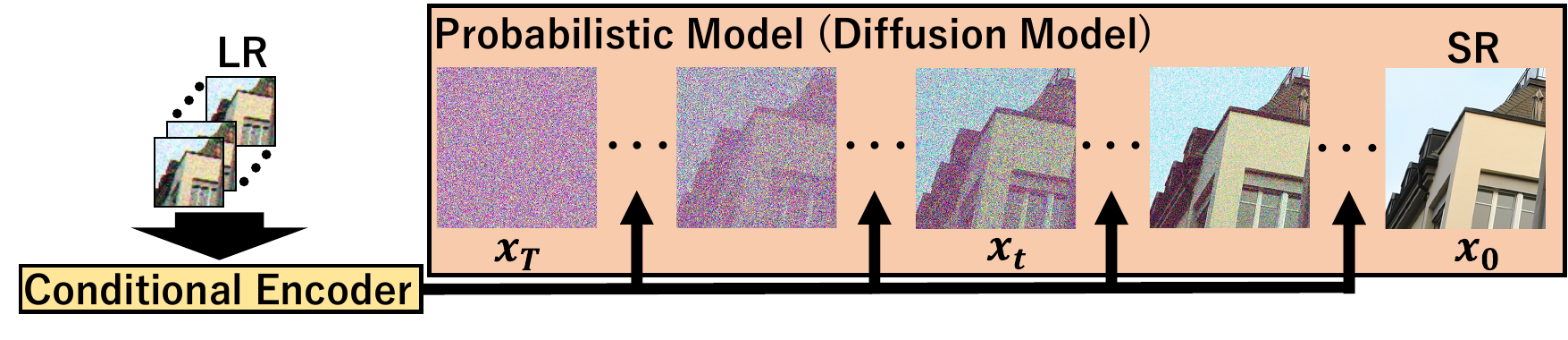}\\
  \vspace*{-2mm}
  (b) Probabilistic models\\
  \includegraphics[width=\linewidth]{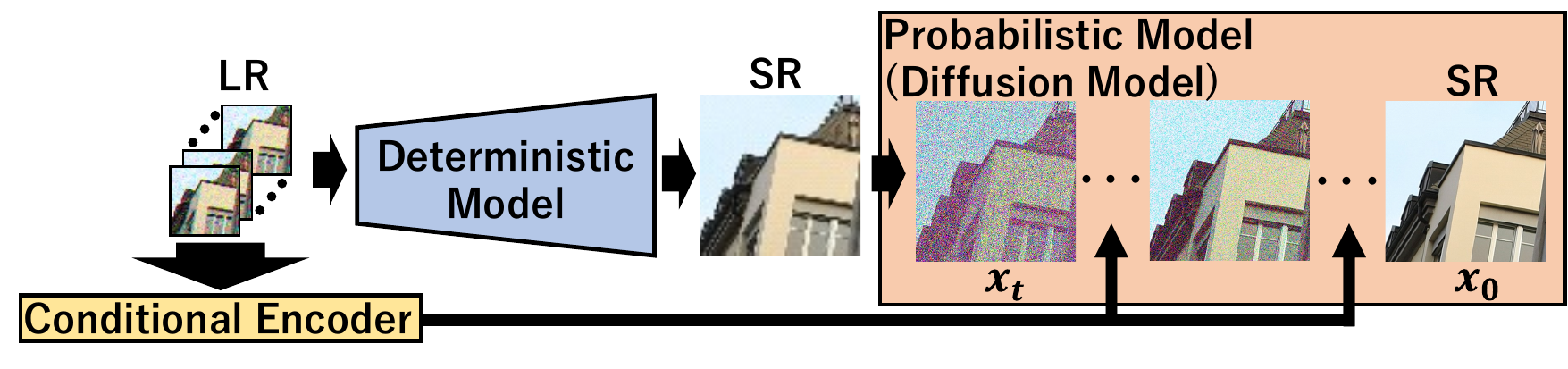}\\
  \vspace*{-2mm}
  (c) Our model (BSRD)
  \vspace*{-2mm}
  \caption{Comparison between prior models and our model (i.e., BSRD).}
  \label{fig:teaser}
  \end{center}
  \vspace*{-4mm}
\end{figure}

As with many computer vision tasks, burst SR can be improved by deep learning (e.g.,~\cite{DBLP:journals/tgrs/MoliniVFM20}).
In previous burst SR methods using deep learning, an SR image is reconstructed in a deterministic manner with a loss function based on the Mean Squared Error (Fig.~\ref{fig:teaser} (a)).
However, such an MSE-based loss function leads to blurred SR images because the MSE is minimized by the average of training HR images~\cite{DBLP:conf/cvpr/LedigTHCCAATTWS17}.

To avoid this problem, this paper aims to improve the perceptual quality of the SR image by probabilistic modeling.
Probabilistic modeling allows us to represent the probabilistic distribution of sharp SR images corresponding to each LR image.
Among such probabilistic models, our method employs diffusion models~\cite{DBLP:conf/nips/HoJA20,DBLP:conf/icml/Sohl-DicksteinW15} (Fig.~\ref{fig:teaser} (b)) because of their performance validated in various computer vision tasks.

While the diffusion model begins with a random noise for image synthesis, the diffusion model should be conditioned for an output image that fits with input images for image enhancement and restoration, including SR.
In SISR using the diffusion model such as~\cite{DBLP:journals/pami/SahariaHCSFN23,DBLP:journals/corr/abs-2305-07015}, an input LR image is used for conditioning the reverse process.
Unlike SISR, however, burst SR accepts multiple displaced LR images.
If such displaced images are directly fed into the diffusion model, the reverse process may reconstruct the blurry SR image.

This paper proposes a burst SR method in which the aforementioned burst images are appropriately employed for improving the perceptual quality of the SR image.
The core contributions of this work are as follows (Fig.~\ref{fig:teaser} (c)):
\begin{itemize}
    \item To use the diffusion model for burst SR, features useful for SR are extracted from input burst images and employed for conditioning the diffusion model.
    \item Since a diffusion model with
    between random noise and output images is not optimized for the SR task, the initial burst SR image reconstructed by simple burst SR is fed into the intermediate step of the diffusion model to focus on refining detailed textures on the burst SR image.
\end{itemize}


\section{Related Work}
\label{section:related}

\subsection{Burst SR}
\label{subsection:burst_SR}

While a RAW image consisting of RGGB channels with high bits per pixel resolution (e.g., 14 bits, 16 bits) is captured by a standard digital camera, the RAW image is converted by an Image Signal Processing (ISP) to its 8-bit RGB image.
General burst SR tasks accept a set of unprocessed RAW images instead of their processed RGB images for the following two reasons.
1) Since the function of ISP depends on the camera, there is a domain gap between the RGB images processed by different cameras~\cite{DBLP:conf/cvpr/BrooksMXCSB19,DBLP:conf/cvpr/IgnatovGT20}.
This domain gap makes it difficult to apply a universal SR model to various camera images.
We expect that the domain gap between RAW images is smaller.
2) Image conversion from a raw image with high bits per pixel to its 8-bit RGB image drops useful cues for burst SR.

Many burst SR methods generally consist of four processes: feature extraction, alignment, fusion, and reconstruction.
The alignment process rectifies image features extracted by the feature extraction process so that the features are spatially consistent.
This spatial rectification is achieved to align all images to the reference frame.
The aligned features are then merged by the fusion process.
The fused features are fed into the reconstruction process to acquire the SR image.

Among the four processes, the alignment process is peculiar and essential to burst SR.
If non-aligned features are directly used for SR, the SR image may be blurred.
While the alignment process is required also for video SR~\cite{DBLP:conf/cvpr/NahTGBHMSL19,DBLP:conf/cvpr/HarisSU19,DBLP:conf/eccv/FuoliHGTREKXLXW20,DBLP:conf/cvpr/HarisSU20}, the properties of displacements in video frames and burst images are different.
Therefore, the alignment process optimized to burst SR is required.
For example, since simple optical flow estimation is employed for the alignment process in DBSR~\cite{DBLP:conf/cvpr/BhatDGT21}, its SR performance is limited because simple optical flow estimation is not optimized for sub-pixel level flows.
In~\cite{DBLP:conf/cvpr/LuoYMLJF0L21,DBLP:conf/cvpr/Guo0MR022}, on the other hand, deformable convolution (DC)~\cite{DBLP:conf/iccv/DaiQXLZHW17} is used for implicit spatial alignment.
Adaptively optimized receptive fields in DC improve the alignment process for burst SR.
The two-step alignment with optical flow estimation and DC is also proposed in~\cite{DBLP:conf/cvpr/LuoLCYWWFSL22}.

However, all burst SR methods introduced in Sec.~\ref{subsection:burst_SR} reconstruct SR images in a deterministic manner, which prones to blurry SR images.

\subsection{Diffusion Models}

A diffusion model consists of the diffusion and reverse processes.
The diffusion process changes an original data (e.g., a noiseless image denoted by $x_{0}$) towards its corresponding random noise (denoted by $x_{T}$) by gradually giving noise through $T$ steps (e.g., $T=1000$ steps).
The reverse process, on the other hand, gradually recovers the original data from a random noise.
Given $x_{t-1}$ at $(t-1)$-th step, $t$-th step in the diffusion process predicts $x_{t}$ according to the conditional probability $q(x_{t}|x_{t-1})$ as follows:
\begin{eqnarray}
    q(x_t|x_{t-1}) &:=& \mathcal{N}(x_t;\sqrt{\alpha_{t}}x_{t-1},\beta_{t}),
    \label{equation:q(x_t|x_{t-1})}
\end{eqnarray}
where $\mathcal{N}(x;\mu,\sigma)$ denotes a Gaussian distribution whose mean and covariance matrix are $\mu$ and $\sigma$, respectively.
$0<\beta_{1}<\beta_{2}<\cdots<\beta_{T}<1$ is a hyper-parameter that adjusts the variance so that $\alpha_{t}:=1-\beta_{t}$.
On the other hand, the reverse process in $t$-th step predicts $x_{t-1}$ according to $p(x_{t-1}|x_{t})$.

Given $x_{t-1}$, small noise, and $x_{t}$ that is $x_{t-1}$ added by small noise, the reverse process described above is trained so that $t$-th step can predict this small noise from $x_{t}$.
The reverse process is generally implemented with U-Net~\cite{DBLP:conf/miccai/RonnebergerFB15}.

As mentioned in Sec.~\ref{section:introduction}, while the reverse process begins from random noise for image synthesis, the reverse process should be controlled to reconstruct an output image that fits with an input image for SR, as described in what follows.

\subsection{Image Conditioning for Diffusion Models}
\label{subsection:condition_diffusion}

\subsubsection{Image Enhancement and Restoration with Diffusion Models}
\label{subsubsection:image_condition_diffusion}

Diffusion models are applied to various image enhancement and restoration tasks, such as image deblurring~\cite{DBLP:conf/cvpr/ChungKKY23,DBLP:conf/icml/MurataSLTUME23}, image inpainting~\cite{DBLP:conf/iclr/ChungKMKY23,DBLP:conf/nips/ChungSRY22}, and image colorization~\cite{DBLP:conf/nips/KawarEES22,DBLP:journals/corr/abs-2211-12343}.
In all of these methods, input degraded images are used to condition the diffusion model for each task.

Unlike these methods that begin with a random noise, 
SDEdit~\cite{DBLP:conf/iclr/MengHSSWZE22} begins from an intermediate step.
SDEdit is designed for realistic image synthesis from sketches.
While the diffusion model is not trained with sketches, its reverse process can convert a sketch to its realistic image by denoising the appropriately-noised sketch from the intermediate step.
While SDEdit is originally proposed for sketch-to-image style transfer, our proposed method applies the SDEdit-style reverse process starting from the intermediate step to burst SR.

\subsubsection{SISR with Diffusion Models}
\label{subsubsection:SISR_diffusion}

As with other image enhancement and restoration tasks mentioned above~\cite{DBLP:conf/cvpr/ChungKKY23,DBLP:conf/icml/MurataSLTUME23,DBLP:conf/iclr/ChungKMKY23,DBLP:conf/nips/ChungSRY22,DBLP:conf/nips/KawarEES22,DBLP:journals/corr/abs-2211-12343}, an input degraded image (i.e., LR image in the SR task) is used for conditioning the diffusion model.
In SR3~\cite{DBLP:journals/pami/SahariaHCSFN23}, the LR image is upscaled by Bicubic interpolation and concatenated with $x_{t}$ for conditioning.
In LDMs~\cite{DBLP:conf/cvpr/RombachBLEO22}, features extracted from the LR image are fed into the middle layers of U-Net in the reverse process through the cross attention mechanism.
As well as the features extracted from the LR image, the number of the diffusion step is also used for step-aware conditioning in Stable SR~\cite{DBLP:journals/corr/abs-2305-07015}.

Unlike these SISR methods~\cite{DBLP:journals/pami/SahariaHCSFN23,DBLP:conf/cvpr/RombachBLEO22,DBLP:journals/corr/abs-2305-07015}, this paper proposes the diffusion model conditioned with burst LR images.


\begin{figure*}[t]
  \begin{center}
  \includegraphics[width=\linewidth]{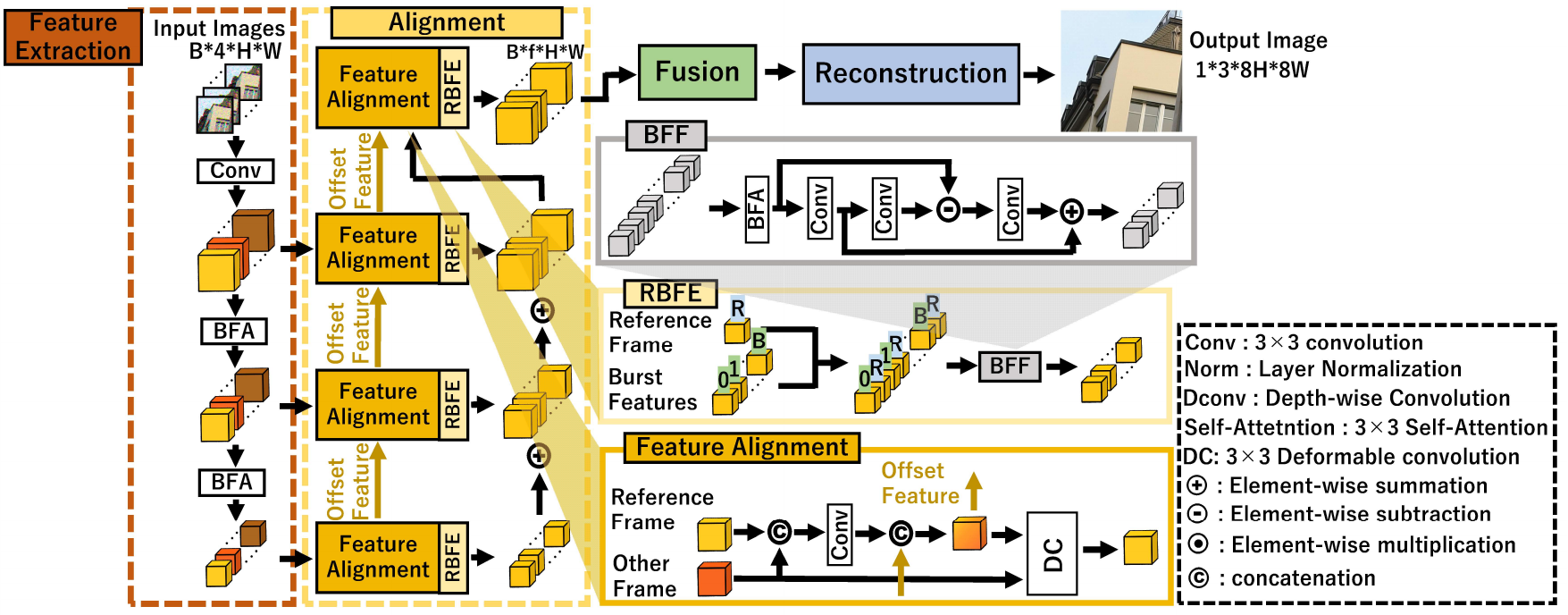}
     \caption{Feature extraction and alignment in Burstormer~\cite{DBLP:conf/cvpr/DudhaneZ0K023}, which are colored by red and yellow, respectively. 
     Different colors in the feature extraction process mean spatial displacements.}
     \label{fig:Burstormer_1}
  \end{center}
\end{figure*}

\begin{figure*}[t]
  \begin{center}
     \includegraphics[width=\linewidth]{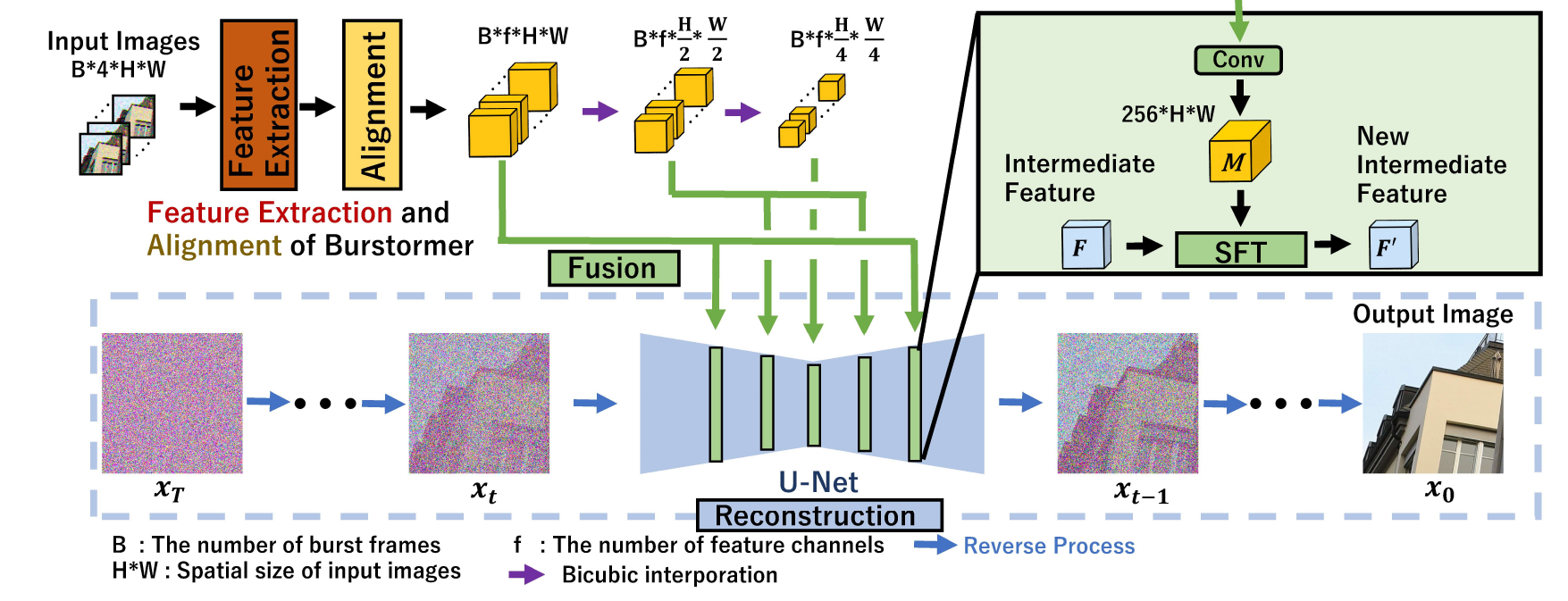}
     \caption{Overview of the feature extraction, alignment, fusion, and reconstruction processes in BSRD. The feature extraction an alignment modules are borrowed from those of Burstormer, which are shown in Fig.~\ref{fig:Burstormer_1}. 
     In our proposed fusion module, SFT~\cite{DBLP:conf/cvpr/WangYDL18} is included in U-Net and used for conditioning with the LR features.
     The reconstruction process is achieved by the reverse process of the diffusion model.}
     \label{fig:BurstSR_Diffusion_Model}
   \end{center}
\end{figure*}

\begin{figure*}[t]
  \begin{center}
     \includegraphics[width=0.9\linewidth]{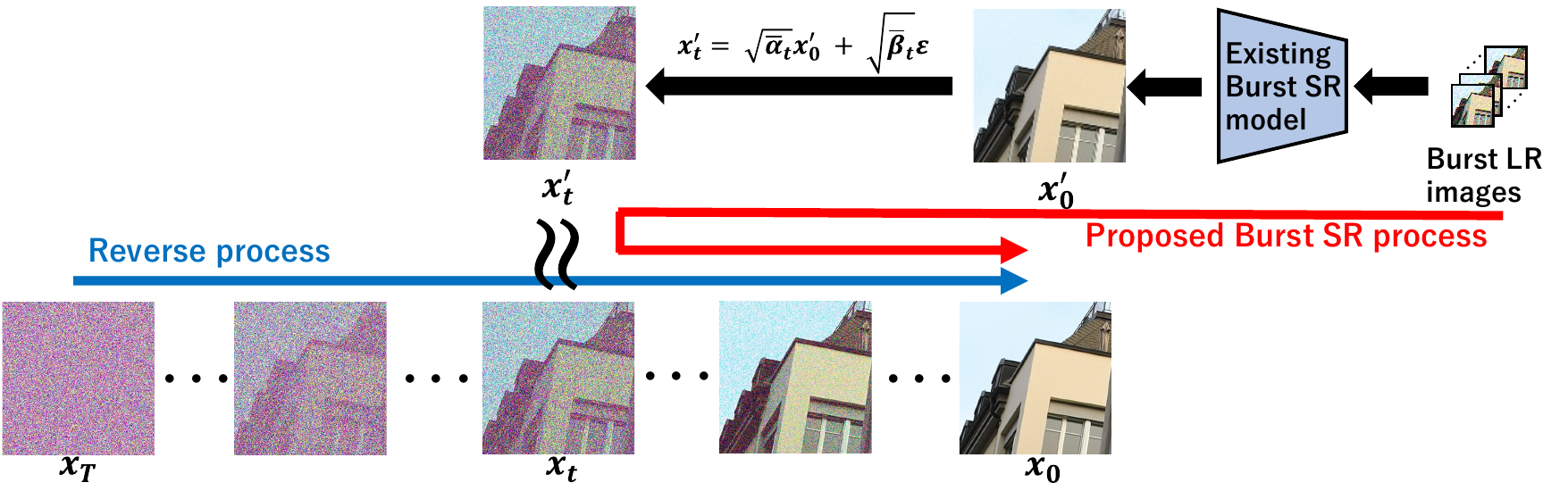}
     \caption{Reverse process from the intermediate step.
     Instead of the reverse process starting from $T$-th step, the initial burst SR image is appropriately noised and fed into the diffusion model from $t$-th step.
     }
     \label{fig:BurstSR_Diffusion_Model_Sample_existing}
   \end{center}
\end{figure*}

\section{Proposed Method}
\label{section:method}

The feature extraction and alignment modules borrowed from the deterministic burst SR method (Burstormer~\cite{DBLP:conf/cvpr/DudhaneZ0K023}) are explained in Sec.~\ref{subsection:burstormer}.
Our probabilistic burst SR method (Burst Super-Resolution with Diffusion Model: BSRD) is proposed in Sec.~\ref{subsubsection:burst_conditioning}.
BSRD is augmented by the reverse process starting from an intermediate step (Sec.~\ref{subsubsection:Sampling_with_LR_and_BSR}).

\subsection{Feature Extraction and Alignment Modules}
\label{subsection:burstormer}

Burstormer~\cite{DBLP:conf/cvpr/DudhaneZ0K023} is one of the state-of-the-art deterministic burst SR methods.
The feature extraction and alignment networks in Burstormer are illustrated in Fig.~\ref{fig:Burstormer_1}.
Given $B$ input burst raw LR images, each of which has four color channels with $W$ and $H$ pixels, features with $B \times f \times W \times H$ channels, where $f$ denotes the feature dimension, are extracted.
Feature channels colored by different colors mean that these channels are spatially displaced. 
These features are fed into the Burst Feature Attention (BFA) modules for noise reduction and downsampling.
Since BFA is implemented by Transformer~\cite{DBLP:conf/cvpr/ZamirA0HK022}, noise is reduced based on the global structure of the images.
These noise reduction and downsampling can simplify feature alignment.
The alignment process consists of the feature alignment and Reference-Based Feature Enrichment (RBFE) modules.
The hierarchically-aligned features are fed into the RBFE modules in each of which the Burst Feature Fusion (BFF) module merges the features of the reference frame and other frames for feature alignment according to the reference frame.

The output features of the four hierarchical feature alignment modules are fed into the fusion module in which the features are merged by average pooling.
The merged features are fed into the reconstruction module for the final SR image.

Since the hierarchical feature representation is powerful for image alignment in burst SR as well as for various computer vision tasks, our method employs the feature extraction and alignment processes used in Burstormer.
On the other hand, the fusion and reconstruction processes are achieved by the diffusion model in our method described in what follows.

\subsection{BSRD: Burst Super-Resolution with Diffusion Model}
\label{subsection:Condition_Feature_Encoder}

\subsubsection{Burst Feature Conditioning for Diffusion Models}
\label{subsubsection:burst_conditioning}

As mentioned in Sec.~\ref{subsection:burstormer}, the feature extraction and alignment modules are borrowed from those of Burstormer, which are shown in Fig.~\ref{fig:Burstormer_1}. 
While the fusion and reconstruction processes in Burstormer are deterministic, BSRD employs the diffusion model for probabilistic burst SR, as illustrated in Fig.~\ref{fig:BurstSR_Diffusion_Model}.

\paragraph{Fusion}

In BSRD, features provided from burst LR images by the feature extraction and alignment modules are used for conditioning the diffusion model to reconstruct an SR image that fits with input burst LR image.
This conditioning should be designed so that the conditioning features are properly formatted according to the network architecture of the reverse process in the diffusion model; inappropriate conditioning decreases the effect of this conditioning.
As well as the reverse process of general diffusion models, that of BSRD is implemented with U-Net.
We assume that hierarchical layers in U-Net should be conditioned for better SR reconstruction.
For conditioning hierarchical layers in U-Net, the output of the alignment module (which is depicted as the orange cuboids with $B \times f \times W \times H$ channels in Fig.~\ref{fig:BurstSR_Diffusion_Model}) is rescaled to the $xy$ dimensions of these hierarchical layers by Bicubic interpolation as depicted by the purple arrows in Fig.~\ref{fig:BurstSR_Diffusion_Model}.
The $B$ channels in each rescaled feature map are merged as done in Burstormer.
For example, the feature map with $B \times f \times W \times H$ channels is converted to the one with $256 \times W \times H$.
The $256$ channels are merged by a convolutional layer.
The merged feature map (denoted by $M$) is fed into U-Net for conditioning.
This conditioning is achieved through Spatial Feature Transformation (SFT)~\cite{DBLP:conf/cvpr/WangYDL18}.
In SFT, the conditioning feature map with $256 \times W \times H$ (i.e., $\bm{M}$ in BSRD) is transformed to two feature maps with $f_{U} \times W \times H$ channels, where $f_{U}$ denotes the feature dimension of each U-Net layer, by two independent convolutional layers.
Given these two feature maps (denoted by $\bm{\alpha}$ and $\bm{\beta}$) and the original feature map in U-Net (denoted by $\bm{F}$ with $f_{U} \times W \times H$ channels), SFT acquires the conditioned feature map denoted by $\bm{F}'$ as follows:
\begin{eqnarray}
    \bm{F'} &=& ( 1 + \bm{\alpha} ) \odot \bm{F} + \bm{\beta}
    \label{equation:SFT}
\end{eqnarray}

The aforementioned fusion process is performed in all steps in the diffusion model.

\paragraph{Reconstruction}

While the reverse process of the diffusion model in BSRD is conditioned as described above, the continuous reverse steps are achieved in BSRD as with in the original diffusion model.
This reverse process is regarded as the reconstruction process in BSRD.

\subsubsection{Efficient and High-quality SR Reconstruction by the Reverse Process from Intermediate Steps}
\label{subsubsection:Sampling_with_LR_and_BSR}

In the reverse process of the diffusion model for image synthesis, it is known that different steps take on different roles~\cite{DBLP:conf/cvpr/ChoiLSKKY22}.
During steps handling noisy images (e.g., at and near $T$-th step), overall color and structure are synthesized.
The overall image appearance is determined during these steps.
In the middle steps, contextual information is reconstructed.
Finally, more detailed textures and boundaries are synthesized during steps handling less-noisy images.
Such an iterative process with many steps requires a huge computational cost, which is one of the major drawbacks of the diffusion model~\cite{DBLP:conf/iclr/SongME21}.
Furthermore, in the reverse process starting from a random sample, the conditioning scheme is required to significantly adjust the reverse process for SR (and other image enhancement and restoration methods) compared with free image synthesis.

To suppress the difficulty in computational cost and SR using the diffusion model, BSRD accepts an initial burst SR image instead of a random noise.
To start the reverse process from the initial burst SR image, it is fed into an intermediate step, instead of $T$-th step, following SDEdit~\cite{DBLP:conf/iclr/MengHSSWZE22}.
With this reverse process from the intermediate step, the number of execution steps is reduced, resulting in reducing the computational cost.
Furthermore, since the difficult steps near $T$-th steps are skipped, the difficulty in reconstructing the SR image close to the input LR image can be reduced, and the reverse process can be trained to focus on reconstructing fine details that are important for the SR task.

The aforementioned reverse process in BSRD is illustrated in Fig.~\ref{fig:BurstSR_Diffusion_Model_Sample_existing}.
Remember $x_{t}$ denotes the image at $t$-th step in the diffusion model.
In BSRD, $x_{\tau}$ is replaced by the initial burst SR image with noise (denoted by $x_{\tau}'$).
This time step $\tau$ is called the reverse start step.
Any SoTA burst SR method can be used for obtaining $x_{\tau}'$.
Let $x_{0}'$ denote the initial burst SR image.
While $x_{0}'$ corresponds to $x_{0}$ with no diffusion noise, $x_{0}'$ may be more blurred than $x_{0}$ because $x_{0}'$ is reconstructed in a deterministic manner.
However, $x_{\tau}'$ computed from $x_{0}'$ must contain the diffusion noise that is included in $x_{\tau}$.
We assume that $x_{\tau}$ can be approximated by $x_{\tau}'$ by giving the diffusion noise at $\tau$-th step to $x_{0}'$.
That is, small differences caused by the blur between $x_{0}$ and $x_{0}'$ can be drowned out by the diffusion noise if $\tau$-th step is sufficiently apart from $0$-th step.
$x_{\tau}'$ is computed from $x_{0}'$ as follows:
\begin{eqnarray}
    x_{\tau}' &=& \sqrt{\smash[b]{\mathstrut\bar{\alpha}_{\tau}}}x_0' + \sqrt{\smash[b]{\bar{\beta}_{\tau}}}\epsilon, \\
    \bar{\alpha}_{\tau} &:=& \prod_{s=1}^{\tau}\alpha_s, \\
    \bar{\beta}_{\tau} &:=& 1 - \bar{\alpha}_{\tau},
  \label{equation:x_t|x_0}
\end{eqnarray}
where $\epsilon$ denotes the zero-mean Gaussian noise with $\sigma=1$.

While the computational cost for computing the initial burst SR image is required, its cost is much smaller than the iterative steps in the diffusion model.

Since BSRD uses the diffusion steps between $\tau$-th and $1$-th steps, $(\tau+1)$-th and greater steps are not required.
Therefore, the diffusion model can be trained only between $\tau$-th and $1$-th steps in the training stage.
This training strategy allows the diffusion model to focus only on reconstructing detailed appearances represented near the $1$-th step, resulting in better SR.


\section{Experimental Results}
\label{section:experiments}

\subsection{Details}

\paragraph{Evaluation Metrics}

As standard evaluation metrics, LPIPS~\cite{DBLP:conf/cvpr/ZhangIESW18}, FID~\cite{DBLP:conf/nips/HeuselRUNH17}, PSNR, and SSIM~\cite{DBLP:journals/tip/WangBSS04} are used.
While PSNR and SSIM are for image distortion-based evaluation, LPIPS and FID are proposed for perceptual quality evaluation.
Lower scores are better in LPIPS and FID, while higher scores are better in PSNR and SSIM.

\paragraph{Training Details}

\begin{figure}
  \begin{center}
     \includegraphics[width=\linewidth]{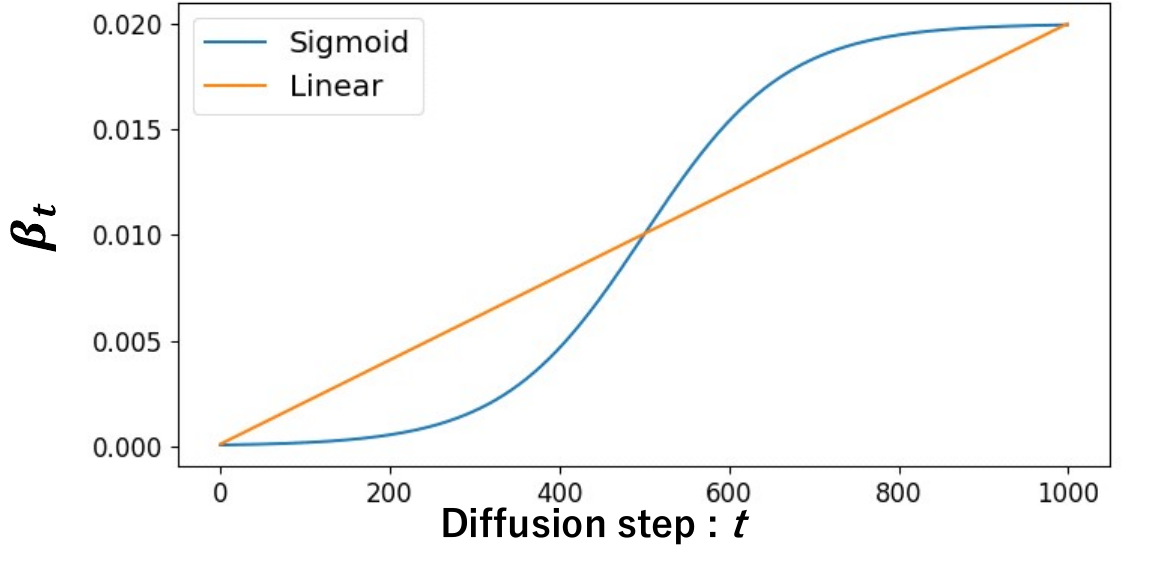}
     \vspace*{-7mm}
     \caption{Comparison between the linear and sigmoid schedulers.}
     \label{fig:graph_betat}
   \end{center}
\end{figure}

The number of diffusion steps is $T = 1000$.
As noise schedulers for the diffusion model, the linear scheduler between $\beta_1 = 10^{-4}$ and $\beta_T = 0.02$ and the Sigmoid scheduler between $\beta_1 = 10^{-5}$ and $\beta_T = 0.02$ are used (Fig.~\ref{fig:graph_betat}).
The mini-batch size is 50.
The optimizer is AdamW~\cite{DBLP:conf/iclr/LoshchilovH19}, and its learning rate is $1 \times 10^{-4}$.

Since our BSRD requires only early steps (i.e., between $1$-th and $\tau$-th steps) in the diffusion model, we can train it only between $1$-th and $\tau$-th steps, as described at the end of Sec.~\ref{subsubsection:Sampling_with_LR_and_BSR}.
For all experiments shown in Sec.~\ref{subsection:synthetic_burst_dataset} and Sec.~\ref{subsection:burst_sr_dataset}, however, the diffusion model was trained between $1$-th and $100$-th steps for simplifying experiments (i.e., for conducting experiments using different $\tau$ with one trained model).
Note that several models evaluated in Sec.~\ref{subsection:detaild_analysis} are trained between different steps to validate the effectiveness of our proposed reverse process from intermediate steps; see Table~\ref{table:result_synth_LR} for details.

\paragraph{Datasets}

The SyntheticBurst and BurstSR datasets are used~\cite{DBLP:conf/cvpr/BhatDGT21}.
On the SyntheticBurst dataset, each sRGB HR image is degraded to its LR image.
This degradation process, including ISP, follows the one proposed in ~\cite{DBLP:conf/cvpr/BhatDGT21} as follows.
Burst LR images, except for the reference frame, are randomly (1) translated up to 24 pixels along $x$ and $y$ axes and (2) rotated up to one degree.
The SyntheticBurst dataset has 46,839 training images and 300 test images.
In the BurstSR dataset, both LR and HR images are real images captured.
Each set of LR images was captured by a burst shot mode of Samsung Galaxy S8.
These LR images are subtly different from each other due to hand shake effects.
Each HR image was captured by CANON 5D Mark IV.
The BurstSR dataset has 5,405 training images and 882 test images.

\paragraph{Burst SR Condition}

In all experiments, the dimensions of LR and HR images are $32 \times 32 \times 4$ RAW images and $256 \times 256 \times 3$ RGB images, respectively.
The number of burst LR images is eight.
The dimension of the conditioning feature is $f=48$.
For comparison, DBSR~\cite{DBLP:conf/cvpr/BhatDGT21}, BIPNet~\cite{DBLP:conf/cvpr/DudhaneZ0K022}, and Burstormer~\cite{DBLP:conf/cvpr/DudhaneZ0K023} are evaluated.
While the authors' pre-trained weights are used for DBSR and BIPNet, the pre-trained weight of Burstormer is further finetuned for better performance.

\begin{figure}[t]
  \begin{center}
     \includegraphics[width=\linewidth]{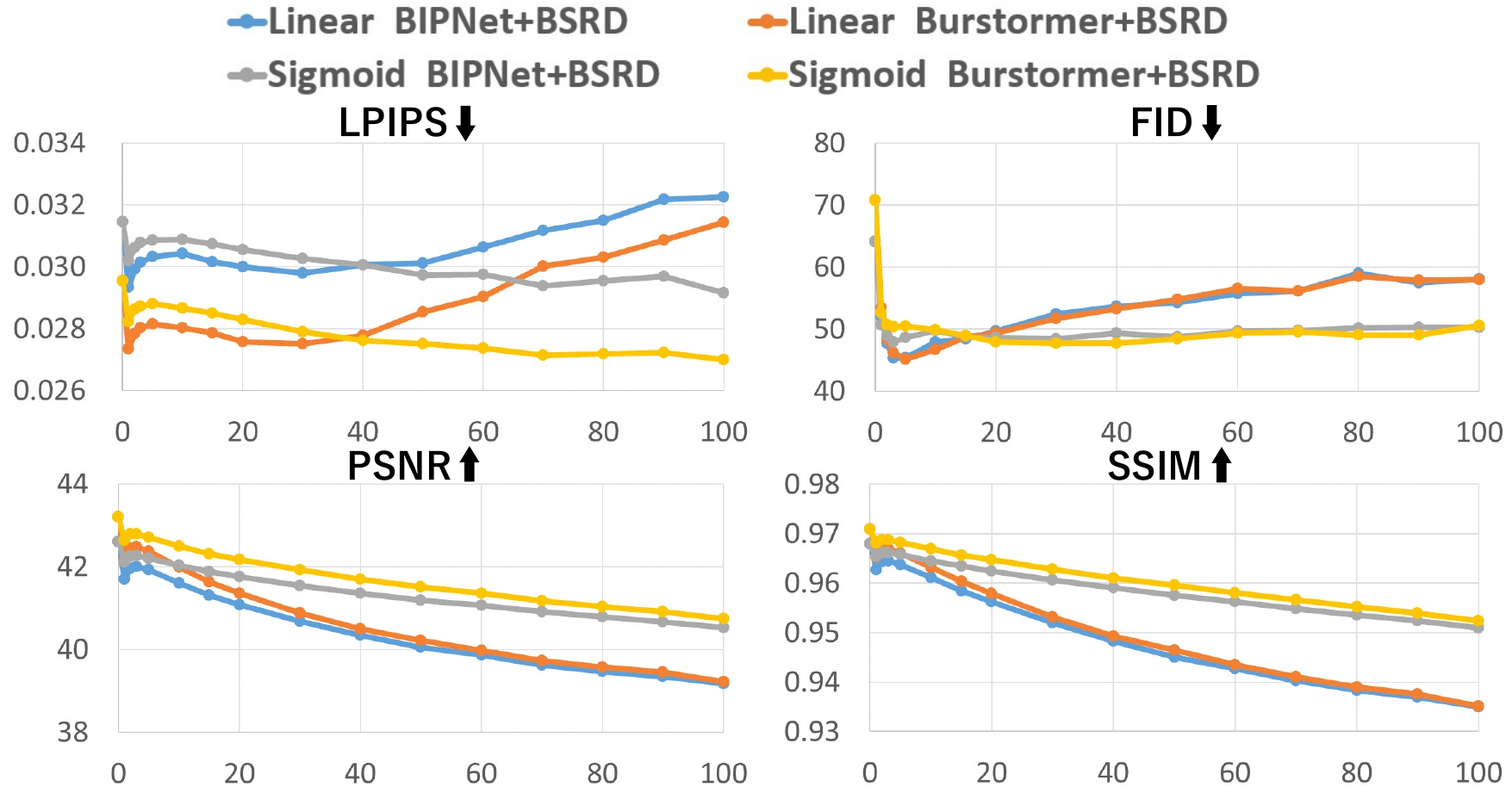}
     \caption{Results on the SyntheticBurst dataset. The vertical and horizontal axes indicate the reverse start step, $\tau$, and the quality measures, respectively.
     While the perceptual scores (i.e., LPIPS and FID) are important for our work, PSNR and SSIM are also shown just for reference.}
     \label{fig:graph_result_synth_exi}
   \end{center}
\end{figure}

Initial burst SR images given to our method (BSRD) are given by BIPNet and Burstormer.
For BSRD, an intermediate step ($\tau$ in Sec.~\ref{subsubsection:Sampling_with_LR_and_BSR}) from which the reverse process starts is an important hyperparameter.
In our experiments, $\tau$ is determined empirically.
Since our goal is to improve the perceptual quality such as LPIPS and FID, $\tau$ giving the best perceptual score in LPIPS and FID is selected.
LPIPS and FID in different steps on the SyntheticBurst dataset are shown in Fig.~\ref{fig:graph_result_synth_exi}.
In the tables in Fig.~\ref{fig:graph_result_synth_exi}, the score at $\tau=0$ equals that of each initial burst SR image.
Based on the results shown in Fig.~\ref{fig:graph_result_synth_exi}, $\tau$ for ``the linear scheduler for LPIPS,'' ``the linear scheduler for FID,'' ``the sigmoid scheduler for LPIPS,'' and ``the sigmoid scheduler for FID'' are 1, 3, 3, and 100, respectively, for BIPNet with BRSD.
The results with these parameters on the SyntheticBurst dataset are shown in Table~\ref{table:result_synth_BSR}.
For Burstomer with BSRD, 1, 5, 30, and 100.
For ``BIPNet'' and ``Burstomer'' with BRSD in the BurstSR dataset, ``6, 30, 60, and 70'' and ``6, 60, 70, and 80,'' respectively, as shown in Table~\ref{table:result_real_BSR}.


\subsection{Results: SyntheticBurst Dataset}
\label{subsection:synthetic_burst_dataset}

\begingroup
\renewcommand{\arraystretch}{1.3}
\begin{table}[t]
    \centering
    \caption{Quantitative comparison results on the SyntheticBurst dataset. The best and second best scores are colored \red{red} and \blue{blue}, respectively.}
    \label{table:result_synth_BSR}
    \begin{adjustbox}{width=\columnwidth,center}
    \begin{tabular}{|l|r|c|llll|}
        \hline
        Methods & \multicolumn{1}{l|}{$\tau$} & \multicolumn{1}{l|}{\begin{tabular}[c]{@{}l@{}}Noise\\ Schedule\end{tabular}} & LPIPS↓ & FID↓ & PSNR↑ & SSIM↑ \\ \hline
        DBSR  & \diagbox[width=9mm,height=3.5mm] & \diagbox[width=14mm,height=3.5mm] & 0.07212 & 99.66 & 40.05 & 0.9482 \\ \hline
        BIPNet  & \diagbox[width=9mm,height=3.5mm] & \diagbox[width=14mm,height=3.5mm] & 0.03145 & 64.17 & \color{blue}42.61 & \color{blue}0.9682 \\ \hline
        \multirow{4}{*}{\textbf{\begin{tabular}[c]{@{}l@{}}BIPNet\\ +BSRD\end{tabular}}}  & 1  & \multirow{2}{*}{linear} & 0.02934 & 51.92 & 41.69 & 0.9627 \\ \cline{2-2} \cline{4-7} 
            & 3 &            & 0.03013 & \color{blue}45.32 & 42.00 & 0.9645 \\ \cline{2-7} 
            & 3 & \multirow{2}{*}{sigmoid} & 0.03077 & 47.88 & 42.26 & 0.9662 \\ \cline{2-2} \cline{4-7} 
            &100&  & 0.02915 & 50.26 & 40.53 & 0.9509  \\ \hline
        Burstormer & \diagbox[width=9mm,height=3mm] &  \diagbox[width=14mm,height=3mm] & 0.02954 & 70.76 & \color{red}43.21 & \color{red}0.9709 \\ \hline
        \multirow{4}{*}{\textbf{\begin{tabular}[c]{@{}l@{}}Burstormer\\ +BSRD\end{tabular}}} & 1 & \multirow{2}{*}{linear} & \color{blue}0.02734 & 53.42 & 42.17 & 0.9653 \\ \cline{2-2} \cline{4-7} 
            & 5 &            & 0.02815 & \color{red}45.16 & 42.37 & 0.9660 \\ \cline{2-7} 
            & 30& \multirow{2}{*}{sigmoid} & 0.02791 & 47.74 & 41.92 & 0.9628 \\ \cline{2-2} \cline{4-7} 
            &100 &                        & \color{red}0.02699 & 50.61 & 40.74 & 0.9524 \\ \hline
    \end{tabular}
    \end{adjustbox}
\end{table}
\endgroup

Quantitative comparison results on the SyntheticBurst dataset are shown in Table~\ref{table:result_synth_BSR}.
Independently of the reverse start step, the noise scheduler, and the base method (i.e., BIPNet or Burstormer), both LPIPS and FID are improved by BSRD from the base method.
BSRD improves LPIPS up to about 8.6\% with the sigmoid scheduler and $\tau=100$ compared with the original Burstormer (i.e., 0.02954 in Burstormer vs. 0.02699 in Burstormer + BSRD).
Furthermore, BSRD improves FID up to about 36\% with the linear scheduler and $\tau=5$ compared with the original Burstormer (i.e., 70.76 in Burstormer vs. 45.16 in Burstormer + BSRD)

However, the performance change depending on $\tau$ differs between the noise schedulers.
With the linear scheduler, both LPIPS and FID degrade from earlier steps (i.e., around 1 and 5 steps in LPIPS and FID, respectively).
With the sigmoid scheduler, on the other hand, these scores improve until later steps.
In particular, LPIPS improves until around 100 steps.
This is because the sigmoid scheduler finely changes the noise at early steps, as shown in Fig.~\ref{fig:graph_betat}.
This scheduling allows the reverse process to focus on detailed textures and boundaries optimized in early steps, as described in Sec.~\ref{subsubsection:Sampling_with_LR_and_BSR}.

Different from the perceptual scores (i.e., LPIPS and FID), the image distortion scores (i.e., PSNR and SSIM) are degraded by our method.
This result is unavoidable due to the tradeoff between the perceptual and image distortion scores, as demonstrated in~\cite{DBLP:conf/cvpr/BlauM18}.

\begin{figure*}
  \begin{center}
     \includegraphics[width=\linewidth]{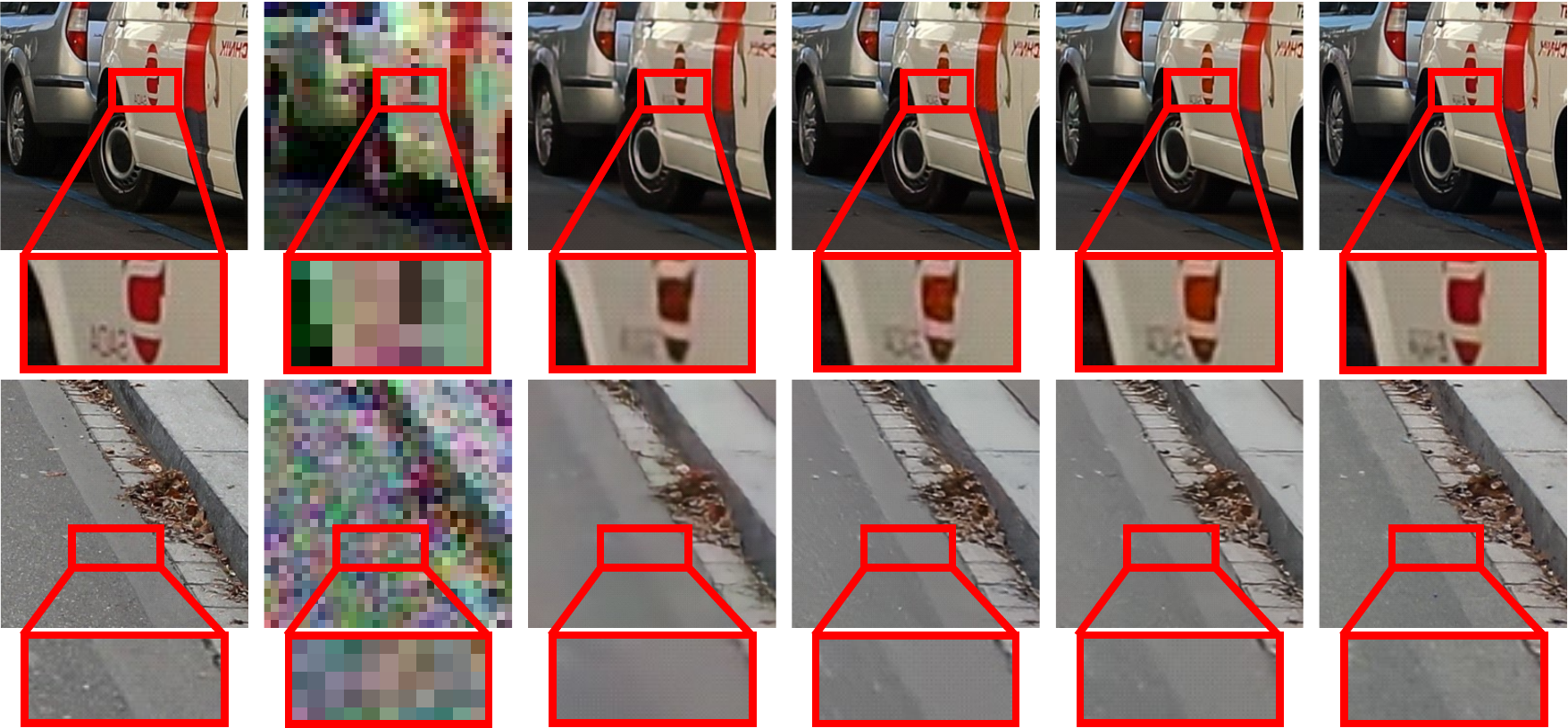}\\
     ~\hspace*{6mm}Ground truth~\hspace*{12mm}
     BurstLR~\hspace*{17mm}
     DBSR~\hspace*{16mm}
     BIPNet~\hspace*{17mm}
     Burstormer~\hspace*{8mm}
     Burstormer+BSRD\\
     \vspace*{-2mm}
     \caption{Visual results on the SyntheticBurst dataset. In both the upper and lower results, a part of each image is zoomed in for detail.
     }
     \label{fig:figure/image_result_synth_exi}
     \vspace*{-3mm}
  \end{center}
\end{figure*}

Visual results are shown in Fig.~\ref{fig:figure/image_result_synth_exi}.
The result of BRSD is obtained with the sigmoid scheduler with $\tau=100$.
In the upper results, while the boundaries between white and red pixels are blurred in the result of Burstormer, our method can reconstruct the sharp boundaries.
In the lower results, the boundary between the different colors on the road is blurred in the result of Burstormer, while our method can reconstruct the clear boundary. 
In addition, the road surface is flat and textureless in the result of Burstormer, while the texture of the road surface is reconstructed by our method.
These results demonstrate that our method can improve the perceptual quality with clear boundary lines.


\subsection{Results: BurstSR dataset}
\label{subsection:burst_sr_dataset}

\begingroup
\renewcommand{\arraystretch}{1.3}
\begin{table}[!t]
    \centering
    \caption[]{Quantitative comparison results on the BurstSR dataset.}
    \begin{center}
    \begin{adjustbox}{width=\columnwidth,center}
    \label{table:result_real_BSR}
    \begin{tabular}{|l|r|c|llll|}
        \hline
        Methods & \multicolumn{1}{l|}{$\tau$} & \multicolumn{1}{l|}{\begin{tabular}[c]{@{}l@{}}Noise\\ Schedule\end{tabular}} & LPIPS↓ & FID↓ & PSNR↑ & SSIM↑  \\ \hline
        DBSR  & \diagbox[width=9mm,height=3.5mm] & \diagbox[width=14mm,height=3.5mm] & 0.05467 & 90.34 & 50.72 & 0.9846  \\ \hline
        BIPNet  & \diagbox[width=9mm,height=3.5mm] & \diagbox[width=14mm,height=3.5mm] & 0.04975 & 75.43 & \color{red}51.55 & \color{red}0.9857 \\ \hline
        \multirow{4}{*}{\textbf{\begin{tabular}[c]{@{}l@{}}BIPNet\\ +BSRD\end{tabular}}}  & 6 & \multirow{2}{*}{linear} & 0.04813 & \color{blue}43.34 & \color{blue}50.91 & \color{blue}0.9850 \\ \cline{2-2} \cline{4-7} 
                                      & 30 &        & \color{red}0.04682 & 46.65 & 50.32 & 0.9836 \\ \cline{2-7} 
                                      & 60 & \multirow{2}{*}{sigmoid} & \color{blue}0.04688 & \color{blue}43.34 & 50.54 & 0.9835 \\ \cline{2-2} \cline{4-7} 
                                      & 70 &  & 0.04719 & \color{red}43.11 & 50.48 & 0.9839  \\ \hline
        Burstormer & \diagbox[width=9mm,height=3.5mm] & \diagbox[width=14mm,height=3.5mm] & 0.05593 & 74.21 & 50.49 & 0.9845\\ \hline
        \multirow{4}{*}{\textbf{\begin{tabular}[c]{@{}l@{}}Burstormer\\ +BSRD\end{tabular}}} & 6 & \multirow{2}{*}{linear} & 0.05445 & 46.85 & 50.11 & 0.9837 \\ \cline{2-2} \cline{4-7} 
                                         & 60 &             & 0.05017 & 48.51 & 49.45 & 0.9815 \\ \cline{2-7} 
                                      & 70 & \multirow{2}{*}{sigmoid} & 0.05238 & 44.54 & 49.94 & 0.9826 \\ \cline{2-2} \cline{4-7} 
                                      & 80 &        & 0.05178 & 45.20 & 49.91 & 0.9823 \\ \hline
    \end{tabular}
    \end{adjustbox}
    \end{center}
\end{table}
\endgroup

Quantitative comparison results on the BurstSR dataset are shown in Table~\ref{table:result_real_BSR}.
As with the SyntheticBurst dataset, the BurstSR dataset demonstrates that our method can improve the perceptual quality.
For example, our method improves LPIPS and FID by 5.8\% and 43\%, respectively (i.e., 0.04975 with BIPNet vs. 0.04642 with our method in LPIPS, and 75.43 with BIPNet vs. 43.11 with our method in FID).
With Burstormer, our method can also improve the perceptual quality scores, while its improvement is lower than that of BIPNet.

\begin{figure*}
  \begin{center}
     \includegraphics[width=\linewidth]{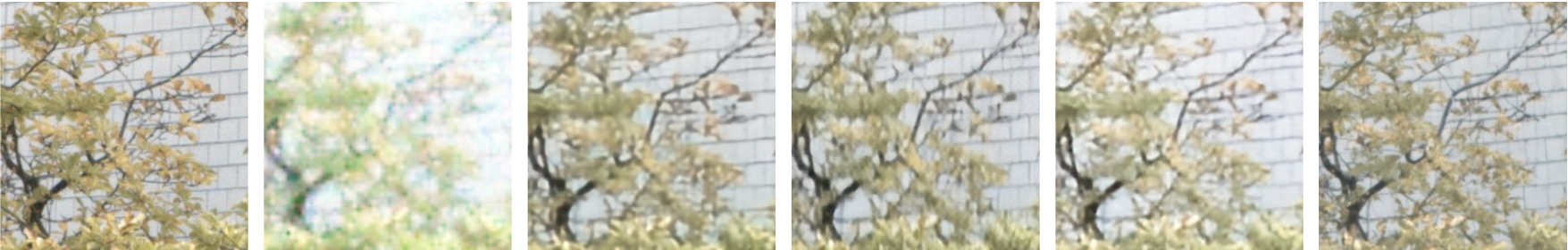}\\
     \includegraphics[width=\linewidth]{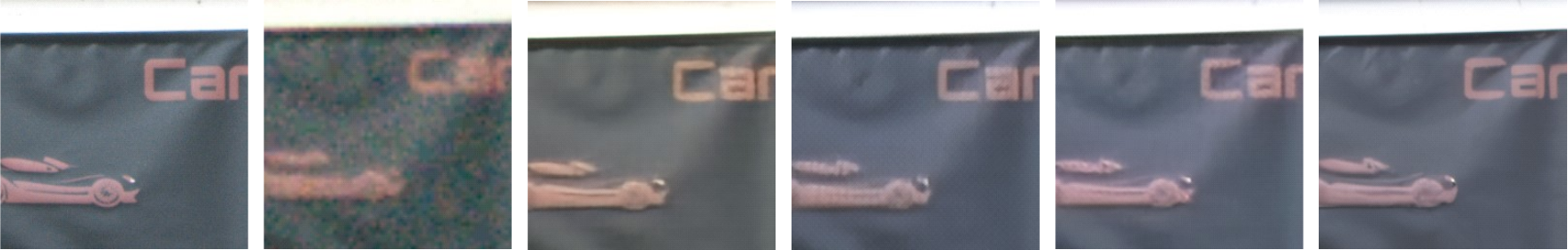}\\
~\hspace*{2mm}Ground truth~\hspace*{12mm}
     BurstLR~\hspace*{16mm}
     DBSR~\hspace*{16mm}
     Burstormer~\hspace*{16mm}
     BIPNet~\hspace*{12mm}
     BIPNet+BSRD\\
     \vspace*{-2mm}
     \caption{Visual results on the BurstSR dataset.
     }
     \label{fig:figure/image_result_real_exi}
     \vspace*{-4mm}
   \end{center}
\end{figure*}

Visual results on the BurstSR dataset are shown in Fig.~\ref{fig:figure/image_result_real_exi}.
Our method (i.e., ``BIPNet+BSRD'' in Fig.~\ref{fig:figure/image_result_real_exi}) is combined with BIPNet because this combination produces the best results on the BurstSR dataset, as shown in Table~\ref{table:result_real_BSR}.
Our method outperforms all other methods, including its base method (i.e., BIPNet), regarding the sharpness of boundary lines.


\subsection{Detailed Analysis}
\label{subsection:detaild_analysis}

\begingroup
\renewcommand{\arraystretch}{1.3}
\begin{table}[!t]
\caption{Detailed analysis on the SyntheticBurst dataset for validating the effectiveness of each component in our BSRD. Results in each row are acquired with the diffusion model trained between $1$-th and $\tau_{L}$-th steps.}
    \begin{center}
    \begin{adjustbox}{width=\columnwidth,center}
        \label{table:result_synth_LR}
        \begin{tabular}{|l|l|r|r|rrrr|}
        \hline
        & Methods  & \multicolumn{1}{l|}{\begin{tabular}[c]{@{}l@{}}$\tau_{L}$\end{tabular}} & \multicolumn{1}{l|}{\begin{tabular}[c]{@{}l@{}}$\tau$\end{tabular}} & \multicolumn{1}{l}{LPIPS↓} & \multicolumn{1}{l}{FID↓} & \multicolumn{1}{l}{PSNR↑} & \multicolumn{1}{l|}{SSIM↑} \\ \hline
        (a) & From random noise  & 1000 & 1000 & 0.06679 & 85.02 & 33.86 & 0.8730 \\ \hline
        (b) & \multirow{2}{*}{\begin{tabular}[c]{@{}l@{}}From bicubic\\ upscaled image\end{tabular}} & 1000 & 400 & 0.07164 & 85.79 & 35.60 & 0.8334 \\ \cline{1-1} \cline{3-8}
                                        (c) & & 100  & 5   & 0.31882 & 206.37 & 31.03 & 0.8178 \\ \hline
        (d) & \multirow{2}{*}{\begin{tabular}[c]{@{}l@{}}From Burstormer\\ SR image\end{tabular}}
                                        & 1000 & 5 & 0.02851 & 49.19 & 42.19 & 0.9649 \\ \cline{1-1} \cline{3-8}
                                        (e) & & 100 & 5 & 0.02815 & 45.16 & 42.37 & 0.9660 \\ \hline
        \end{tabular}
    \end{adjustbox}
    \end{center}
\end{table}
\endgroup

To validate the effectiveness of each implementation in our proposed BSRD, the following experiments are conducted:
\begin{itemize}
    \item (a) begins from random noise (i.e., $\tau=1000$) instead of from the intermediate step.
    \item (b) and (c) use the initial image upscaled by bicubic interpolation instead of the burst SR image reconstructed by Burstormer. 
    \item (d) and (e) use the burst SR image reconstructed by Burstormer. (e) is our proposed implementation of BSRD.
\end{itemize}
We can see the following observations in Table~\ref{table:result_synth_LR}:
\begin{enumerate}
    \item In comparison between (a) and (d,e), the reverse process from the intermediate step is better.
    \item In comparison between (b,c) and (d,e), the reverse process with the initial burst SR image is better.
    \item In comparison between (d) and (e), the model trained only from $\tau$-th step is better.
\end{enumerate}


\section{Concluding Remarks}
\label{section:conclusion}

This paper proposed a combination of burst SR and diffusion models for improving the perceptual quality of burst SR.
Unlike prior diffusion-based burst SR methods, which start from random noise, our BSRD starts the reverse process of the diffusion model from an intermediate step.
This allows the diffusion model to efficiently focus on reconstructing detailed textures and boundaries in earlier diffusion steps.
Furthermore, hierarchical burst SR features are employed to condition the reverse process to optimize it for the burst SR task.

Future work includes more efficiency improvement.
It is validated that latent diffusion is useful for SISR as well as for other computer vision tasks~\cite{DBLP:conf/cvpr/RombachBLEO22,CVPR2024READ}.
For burst SR also, the effectiveness of the latent diffusion should be explored.

This work is partly supported by JSPS KAKENHI 22H03618.


\bibliographystyle{unsrt}
\bibliography{main}

\begin{thebibliography}{10}

\bibitem{DBLP:journals/pami/DongLHT16}
Chao Dong, Chen~Change Loy, Kaiming He, and Xiaoou Tang.
\newblock Image super-resolution using deep convolutional networks.
\newblock {\em {IEEE} Trans. Pattern Anal. Mach. Intell.}, 38(2):295--307, 2016.

\bibitem{ntire2018}
Radu Timofte et~al.
\newblock Ntire 2018 challenge on single image super-resolution: Methods and results.
\newblock In {\em CVPRW}, 2018.

\bibitem{DBLP:conf/cvpr/HarisSU18}
Muhammad Haris, Gregory Shakhnarovich, and Norimichi Ukita.
\newblock Deep back-projection networks for super-resolution.
\newblock In {\em CVPR}, 2018.

\bibitem{DBLP:conf/iccvw/GuLZXYZYSTDLDLG19}
Shuhang Gu et~al.
\newblock {AIM} 2019 challenge on image extreme super-resolution: Methods and results.
\newblock In {\em ICCVW}, 2019.

\bibitem{DBLP:journals/pami/HarisSU21}
Muhammad Haris, Greg Shakhnarovich, and Norimichi Ukita.
\newblock Deep back-projectinetworks for single image super-resolution.
\newblock {\em {IEEE} Trans. Pattern Anal. Mach. Intell.}, 43(12):4323--4337, 2021.

\bibitem{DBLP:conf/cvpr/BhatDGT21}
Goutam Bhat, Martin Danelljan, Luc~Van Gool, and Radu Timofte.
\newblock Deep burst super-resolution.
\newblock In {\em CVPR}, 2021.

\bibitem{DBLP:journals/corr/abs-2106-03839}
Goutam Bhat et~al.
\newblock {NTIRE} 2021 challenge on burst super-resolution: Methods and results.
\newblock In {\em CVPRW}, 2021.

\bibitem{DBLP:conf/cvpr/BhatDTCCCCCDFGG22}
Goutam Bhat et~al.
\newblock {NTIRE} 2022 burst super-resolution challenge.
\newblock In {\em CVPRW}, 2022.

\bibitem{DBLP:journals/tgrs/MoliniVFM20}
Andrea~Bordone Molini, Diego Valsesia, Giulia Fracastoro, and Enrico Magli.
\newblock Deepsum: Deep neural network for super-resolution of unregistered multitemporal images.
\newblock {\em {IEEE} Trans. Geosci. Remote. Sens.}, 58(5):3644--3656, 2020.

\bibitem{DBLP:conf/cvpr/LedigTHCCAATTWS17}
Christian Ledig, Lucas Theis, Ferenc Huszar, Jose Caballero, Andrew Cunningham, Alejandro Acosta, Andrew~P. Aitken, Alykhan Tejani, Johannes Totz, Zehan Wang, and Wenzhe Shi.
\newblock Photo-realistic single image super-resolution using a generative adversarial network.
\newblock In {\em CVPR}, 2017.

\bibitem{DBLP:conf/nips/HoJA20}
Jonathan Ho, Ajay Jain, and Pieter Abbeel.
\newblock Denoising diffusion probabilistic models.
\newblock In {\em NeurIPS}, 2020.

\bibitem{DBLP:conf/icml/Sohl-DicksteinW15}
Jascha Sohl{-}Dickstein, Eric~A. Weiss, Niru Maheswaranathan, and Surya Ganguli.
\newblock Deep unsupervised learning using nonequilibrium thermodynamics.
\newblock In {\em ICML}, 2015.

\bibitem{DBLP:journals/pami/SahariaHCSFN23}
Chitwan Saharia, Jonathan Ho, William Chan, Tim Salimans, David~J. Fleet, and Mohammad Norouzi.
\newblock Image super-resolution via iterative refinement.
\newblock {\em {IEEE} Trans. Pattern Anal. Mach. Intell.}, 45(4):4713--4726, 2023.

\bibitem{DBLP:journals/corr/abs-2305-07015}
Jianyi Wang, Zongsheng Yue, Shangchen Zhou, Kelvin C.~K. Chan, and Chen~Change Loy.
\newblock Exploiting diffusion prior for real-world image super-resolution.
\newblock {\em CoRR}, abs/2305.07015, 2023.

\bibitem{DBLP:conf/cvpr/BrooksMXCSB19}
Tim Brooks, Ben Mildenhall, Tianfan Xue, Jiawen Chen, Dillon Sharlet, and Jonathan~T. Barron.
\newblock Unprocessing images for learned raw denoising.
\newblock In {\em CVPR}, 2019.

\bibitem{DBLP:conf/cvpr/IgnatovGT20}
Andrey Ignatov, Luc~Van Gool, and Radu Timofte.
\newblock Replacing mobile camera {ISP} with a single deep learning model.
\newblock In {\em {CVPR} Workshops}, 2020.

\bibitem{DBLP:conf/cvpr/NahTGBHMSL19}
Seungjun Nah et~al.
\newblock {NTIRE} 2019 challenge on video super-resolution: Methods and results.
\newblock In {\em {CVPR} Workshop}, 2019.

\bibitem{DBLP:conf/cvpr/HarisSU19}
Muhammad Haris, Gregory Shakhnarovich, and Norimichi Ukita.
\newblock Recurrent back-projection network for video super-resolution.
\newblock In {\em CVPR}, 2019.

\bibitem{DBLP:conf/eccv/FuoliHGTREKXLXW20}
Dario Fuoli et~al.
\newblock {AIM} 2020 challenge on video extreme super-resolution: Methods and results.
\newblock In {\em {ECCV} Workshop}, 2020.

\bibitem{DBLP:conf/cvpr/HarisSU20}
Muhammad Haris, Greg Shakhnarovich, and Norimichi Ukita.
\newblock Space-time-aware multi-resolution video enhancement.
\newblock In {\em CVPR}, 2020.

\bibitem{DBLP:conf/cvpr/LuoYMLJF0L21}
Ziwei Luo, Lei Yu, Xuan Mo, Youwei Li, Lanpeng Jia, Haoqiang Fan, Jian Sun, and Shuaicheng Liu.
\newblock {EBSR:} feature enhanced burst super-resolution with deformable alignment.
\newblock In {\em {CVPR} Workshops}, 2021.

\bibitem{DBLP:conf/cvpr/Guo0MR022}
Shi Guo, Xi~Yang, Jianqi Ma, Gaofeng Ren, and Lei Zhang.
\newblock A differentiable two-stage alignment scheme for burst image reconstruction with large shift.
\newblock In {\em CVPR}, 2022.

\bibitem{DBLP:conf/iccv/DaiQXLZHW17}
Jifeng Dai, Haozhi Qi, Yuwen Xiong, Yi~Li, Guodong Zhang, Han Hu, and Yichen Wei.
\newblock Deformable convolutional networks.
\newblock In {\em ICCV}, 2017.

\bibitem{DBLP:conf/cvpr/LuoLCYWWFSL22}
Ziwei Luo, Youwei Li, Shen Cheng, Lei Yu, Qi~Wu, Zhihong Wen, Haoqiang Fan, Jian Sun, and Shuaicheng Liu.
\newblock {BSRT:} improving burst super-resolution with swin transformer and flow-guided deformable alignment.
\newblock In {\em {CVPR} Workshops}, 2022.

\bibitem{DBLP:conf/miccai/RonnebergerFB15}
Olaf Ronneberger, Philipp Fischer, and Thomas Brox.
\newblock U-net: Convolutional networks for biomedical image segmentation.
\newblock In {\em MICCAI}, 2015.

\bibitem{DBLP:conf/cvpr/ChungKKY23}
Hyungjin Chung, Jeongsol Kim, Sehui Kim, and Jong~Chul Ye.
\newblock Parallel diffusion models of operator and image for blind inverse problems.
\newblock In {\em CVPR}, 2023.

\bibitem{DBLP:conf/icml/MurataSLTUME23}
Naoki Murata, Koichi Saito, Chieh{-}Hsin Lai, Yuhta Takida, Toshimitsu Uesaka, Yuki Mitsufuji, and Stefano Ermon.
\newblock Gibbsddrm: {A} partially collapsed gibbs sampler for solving blind inverse problems with denoising diffusion restoration.
\newblock In {\em ICML}, 2023.

\bibitem{DBLP:conf/iclr/ChungKMKY23}
Hyungjin Chung, Jeongsol Kim, Michael~Thompson McCann, Marc~Louis Klasky, and Jong~Chul Ye.
\newblock Diffusion posterior sampling for general noisy inverse problems.
\newblock In {\em ICLR}, 2023.

\bibitem{DBLP:conf/nips/ChungSRY22}
Hyungjin Chung, Byeongsu Sim, Dohoon Ryu, and Jong~Chul Ye.
\newblock Improving diffusion models for inverse problems using manifold constraints.
\newblock In {\em NeurIPS}, 2022.

\bibitem{DBLP:conf/nips/KawarEES22}
Bahjat Kawar, Michael Elad, Stefano Ermon, and Jiaming Song.
\newblock Denoising diffusion restoration models.
\newblock In {\em NeurIPS}, 2022.

\bibitem{DBLP:journals/corr/abs-2211-12343}
Xiangming Meng and Yoshiyuki Kabashima.
\newblock Diffusion model based posterior sampling for noisy linear inverse problems.
\newblock {\em CoRR}, abs/2211.12343, 2022.

\bibitem{DBLP:conf/iclr/MengHSSWZE22}
Chenlin Meng, Yutong He, Yang Song, Jiaming Song, Jiajun Wu, Jun{-}Yan Zhu, and Stefano Ermon.
\newblock Sdedit: Guided image synthesis and editing with stochastic differential equations.
\newblock In {\em ICLR}, 2022.

\bibitem{DBLP:conf/cvpr/RombachBLEO22}
Robin Rombach, Andreas Blattmann, Dominik Lorenz, Patrick Esser, and Bj{\"{o}}rn Ommer.
\newblock High-resolution image synthesis with latent diffusion models.
\newblock In {\em CVPR}, 2022.

\bibitem{DBLP:conf/cvpr/DudhaneZ0K023}
Akshay Dudhane, Syed~Waqas Zamir, Salman Khan, Fahad~Shahbaz Khan, and Ming{-}Hsuan Yang.
\newblock Burstormer: Burst image restoration and enhancement transformer.
\newblock In {\em CVPR}, 2023.

\bibitem{DBLP:conf/cvpr/WangYDL18}
Xintao Wang, Ke~Yu, Chao Dong, and Chen~Change Loy.
\newblock Recovering realistic texture in image super-resolution by deep spatial feature transform.
\newblock In {\em CVPR}, 2018.

\bibitem{DBLP:conf/cvpr/ZamirA0HK022}
Syed~Waqas Zamir, Aditya Arora, Salman Khan, Munawar Hayat, Fahad~Shahbaz Khan, and Ming{-}Hsuan Yang.
\newblock Restormer: Efficient transformer for high-resolution image restoration.
\newblock In {\em CVPR}, 2022.

\bibitem{DBLP:conf/cvpr/ChoiLSKKY22}
Jooyoung Choi, Jungbeom Lee, Chaehun Shin, Sungwon Kim, Hyunwoo Kim, and Sungroh Yoon.
\newblock Perception prioritized training of diffusion models.
\newblock In {\em CVPR}, 2022.

\bibitem{DBLP:conf/iclr/SongME21}
Jiaming Song, Chenlin Meng, and Stefano Ermon.
\newblock Denoising diffusion implicit models.
\newblock In {\em ICLR}, 2021.

\bibitem{DBLP:conf/cvpr/ZhangIESW18}
Richard Zhang, Phillip Isola, Alexei~A. Efros, Eli Shechtman, and Oliver Wang.
\newblock The unreasonable effectiveness of deep features as a perceptual metric.
\newblock In {\em CVPR}, 2018.

\bibitem{DBLP:conf/nips/HeuselRUNH17}
Martin Heusel, Hubert Ramsauer, Thomas Unterthiner, Bernhard Nessler, and Sepp Hochreiter.
\newblock Gans trained by a two time-scale update rule converge to a local nash equilibrium.
\newblock In {\em NIPS}, 2017.

\bibitem{DBLP:journals/tip/WangBSS04}
Zhou Wang, Alan~C. Bovik, Hamid~R. Sheikh, and Eero~P. Simoncelli.
\newblock Image quality assessment: from error visibility to structural similarity.
\newblock {\em {IEEE} Trans. Image Process.}, 13(4):600--612, 2004.

\bibitem{DBLP:conf/iclr/LoshchilovH19}
Ilya Loshchilov and Frank Hutter.
\newblock Decoupled weight decay regularization.
\newblock In {\em ICLR}, 2019.

\bibitem{DBLP:conf/cvpr/DudhaneZ0K022}
Akshay Dudhane, Syed~Waqas Zamir, Salman Khan, Fahad~Shahbaz Khan, and Ming{-}Hsuan Yang.
\newblock Burst image restoration and enhancement.
\newblock In {\em CVPR}, 2022.

\bibitem{DBLP:conf/cvpr/BlauM18}
Yochai Blau and Tomer Michaeli.
\newblock The perception-distortion tradeoff.
\newblock In {\em CVPR}, 2018.

\bibitem{CVPR2024READ}
Takeru Oba, Matthew Walter, and Norimichi Ukita.
\newblock Read: Retrieval-enhanced asymmetric diffusion for motion planning.
\newblock In {\em CVPR}, 2024.

\end{thebibliography}

\end{document}